\documentclass[acmsmall]{acmart}

\AtBeginDocument{%
  \providecommand\BibTeX{{%
    \normalfont B\kern-0.5em{\scshape i\kern-0.25em b}\kern-0.8em\TeX}}}

\begin{document}

\title{Fine-tuning Transformer-based Encoder for Turkish Language Understanding Tasks}

\author{Savaş Yıldırım}
\email{savas.yildirim@bilgi.edu.tr}
\orcid{0000-0002-7764-2891}
\affiliation{%
  \institution{Istanbul Bilgi University, Faculty of Engineering and Natural Sciences}
  \streetaddress{Eski Silahtarağa Elektrik Santralı Kazım Karabekir Cad. No: 2/13}
  \city{Istanbul}
  \state{Turkey}
  \postcode{34060}
}

\renewcommand{\shortauthors}{Savaş Yıldırım}

\begin{abstract}
  Deep learning-based and lately Transformer-based language models have been dominating the studies of natural language processing in the last years. Thanks to their accurate and fast fine-tuning characteristics, they have outperformed traditional machine learning-based approaches and achieved state-of-the-art results for many challenging natural language understanding (NLU) problems. Recent studies showed that the Transformer-based models such as BERT, which is Bidirectional Encoder Representations from Transformers, have reached impressive achievements on many tasks. Moreover, thanks to their transfer learning capacity, these architectures allow us to transfer pre-built models and fine-tune them to specific NLU tasks such as question answering. In this study, we provide a Transformer-based model and a baseline benchmark for the Turkish Language. We successfully fine-tuned a Turkish BERT model, namely BERTurk that is trained with base settings, to many downstream tasks and evaluated with a the Turkish Benchmark dataset. We showed that our studies significantly outperformed other existing baseline approaches for Named-Entity Recognition, Sentiment Analysis, Question Answering and Text Classification in Turkish Language. We publicly released these four fine-tuned models and resources in reproducibility and with the view of supporting other Turkish researchers and applications.
\end{abstract}



\maketitle

\section{Introduction}

The advent of deep learning and transfer learning in the field of natural language processing (NLP) has achieved significant improvements for many tasks. Sequence transduction models such as RNN and LSTM are based on complex recurrent neural networks. They successfully connected the encoder and decoder for sequence to sequence problem \cite{Zhang2020}. However one of the computational bottlenecks of RNN-based models is processing very long sequences and they have difficulty in keeping long-distance dependencies.

The Transformers \cite{Vaswani17} overcome this limitation and simplify encoder-decoder network architecture by employing attention mechanisms, dropping recurrence or convolutions entirely. The studies showed that the Transformer could be trained significantly faster than other standard RNN architectures since that RNN models are based on more complex recurrent layers. The initial experiments on some machine translation tasks indicated that the Transformer-based models are superior in quality. They are more parallelizable and further requiring significantly less time to train. Finally, the Transformer-based models became a new paradigm in NLP and led to many successful derivatives that provides state-of-the-art general-purpose architectures tested on General Language Understanding Evaluation (GLUE) Benchmark tasks \cite{glue}; BERT \cite{bert}, GPT-2 \cite{gpt}, RoBERTa \cite{roberta}, XLM \cite{xlm}, XLNet \cite{xlnet}, T5 \cite{t5} and so forth. 

Main advantage of the transformer models is to provide fast and cheap fine-tuning in terms of space and time complexity by transferring the base language models trained in the pre-training phase. While the pre-training phase demands huge amounts of data, the fine-tuning phase requires relatively small amounts of data depending on downstream tasks. The pre-training phase trains the language model so that it can build internal representations such as contextual word embeddings or sentence encoding  \cite{elmo}, which allows there after a model to be reused for different downstream tasks even with little amount of labeled data. The contextualized word embeddings \cite{elmo}, addressed the well known word sense problem and represent the word within its context contrary to other traditional word embeddings like word2vec \cite{mikolov}, GloVe \cite{glove}, fastText\cite{fasttext}. Another advantage of the attention mechanism allows the model to better resolve long-term dependencies and coreference relation between entities \cite{joshi}. This property is important especially for machine translation and question answering problems.

In this study we provide a benchmark for Natural Language Understanding (NLU) downstreams tasks for the Turkish Language. We fine-tuned a pre-built BERT language model, BERTurk \cite{berturk}, to four downstream tasks; \textit{Named-Entity Recognition, Sentiment Analysis, Question Answering and Text Classification}. Compared to the baseline approaches, we have got successful results in the NLU tasks. In  order to reproduce the result and support the community, we publicly released these four fine-tuned models. This study is considered as the first successful attempt to fine-tune the BERT model for these four tasks in the Turkish Language.

\section{Related Works for Language Models}

The traditional language models have been based on n-grams and they were simply a probability distribution over word sequences with an objective function of predicting the last \textit{n-th word} from previous \textit{(n-1) words}. One drawback of n-gram language model is that as the model is being trained with a very large corpus, the number of unique words increases and therefore sparse data problem arises \cite{Zhang2020}. 

Modern neural networks sorted the dimensionality problem out by using continuous and dense representation for the words, called \textit{word embeddings}. The early neural language representation employed encoders to produce static word embeddings. The most popular models, such as \textit{word2vec \cite{mikolov}, glove \cite{glove} and fastText \cite{fasttext}} transformed unsupervised textual data into a supervised one by either predicting the target word using context or predicting contextual words by the target word based on a sliding window of n-gram context. The final output turns out to be context-independent word embeddings which ignores word sense.

The main problem of these early models is that they never use the sentence level representation. Second problem is that the senses of the words as one of the oldest problems in computational linguistics are ignored and a single fixed representation is assigned to each word or token. Further studies used a neural network component which encodes the complete sentence-level information and \textit{contextual word embeddings} which finally alleviate word sense problems. In early days, the essential architectures such as RNN, CNN, LSTM have been widely used as encoder and decoder in a sequence to sequence problems.

\textit{ELMo (Embeddings from Language Models)} representations \cite{elmo}, which is based on LSTM, successfully built \textit{deep contextualized word representations} and solved the sense and sentence encoding problem. The ELMo representations differ from traditional word embeddings in that each word representation is a function of the entire input sentence.

Later on, the Transformers \cite{Vaswani17} simplified sequence-to-sequence models by removing recurrency and only keeping attention. The Transformer-based architecture such as GPT \cite{gpt} and BERT \cite{bert} successfully captured the contextualized sentence-level language representations and word embeddings in a modern way. These sentence encoders differ from sliding window approaches in that they read a complete sentence instead of any fixed sized word sequences defined by a sliding window. While Elmo used bidirectional LM to read the entire sentence independently in both directions and concatenates trained forward and backward LSTMs, BERT adopted Transformer to read and process the entire input at once in a simultaneous bidirectional way.

BERT algorithm is able to read the input in a bidirectional way rather than unidirectional. The BERT model trains two different objectives simultaneously; \textit{Masked Language Model (MLM)} and \textit{ Next Sentence Prediction (NSP)}. MLM is based on optimizing the prediction of the masked words in sentences that are randomly chosen. MLM randomly replaces some masked tokens with a special symbol, such as \textbf{\textit{[MASK]}}. Then, the model training objective is to predict the masked symbol based on the non-masked context. The major difference between Transformer-based BERT and LSMT-based ELMo is that the predictions in BERT are based on the entire context rather than only one direction. The BERT is however criticized that the masks could be never seen at fine-tuning, which faces a mismatch between pre-training and fine-tuning phases. NSP is based on a model that detects whether two sentences follow each other or not. Some other transformer models such as RoBERTa \cite{roberta} or Albert \cite{albert} employ different approaches for the objective functions to improve the results.


Although both RNN and Transformer architectures gained spectacular achievements, the main limitation is said to be understanding long-term dependencies. Transformer-XL \cite{txl} put together the advantages of these two architectures. It employs the self-attention modules on each contextual segment of input and a RNN mechanism to learn dependencies between successive segments. Based on the inspirations of impressive RNN and Transformer-based models like ELMo \cite{elmo} and BERT \cite{bert}, recently a wide range of LM derivatives have been proposed. The variants such as \textit{ Electra \cite{electra}, Albert \cite{albert}, T5 \cite{t5}, RoBERTa \cite{roberta}} are gaining successful achievements on GLUE tasks \cite{glue} which is a set of tools for evaluating the performance of models across a diverse collection of common NLU tasks.

\section{The Transformer-based Model for the Turkish Language}

The advantage of the Transformers is that the same architecture could be used in both pre-training and fine-tuning except output layers. The pre-trained model parameters are used to initialize models for different downstream tasks. Pre-training and fine-tuning made it possible to apply transfer learning in NLP and have become a new paradigm. In this section, we describe how to fine-tune BERT-based models for a variety of problems in Turkish; \textit{Question Answering, Sentiment Analysis, Named-Entity Recognition and Text classification}. We also discuss how to evaluate them with a set of benchmark datasets.  

\subsection{Fine-tuning BERT model for Turkish}

We fine-tuned four NLU downstream tasks based on \textit{BERTurk} \cite{berturk} and uploaded to the repository. BERTurk \cite{berturk} has been successfully pre-trained as a Transformer model with a huge Turkish corpus and shared with the NLP community. There are two alternatives for BERT training; \textit{base and large}. While base architecture has \textit{(L=12, H=768, A=12, Total Parameters=110M)}, large one has \textit{(L=24, H=1024, A=16, Total Parameters=340M)} where L is number of layers, and H is hidden size, A is number of self attention heads.
The model can be used \textit{cased} or \textit{uncased} as a way of formatting input text. For the Turkish language, we have currently only base pre-trained LM of BERT with cased or uncased. 

The last version of Turkish BERT model, which we used in this study, has been trained on a huge training corpus with file size of 35GB and 45B tokens. The compiled corpus consists of online resources such as \textit{OSCAR Corpus} \footnote{https://traces1.inria.fr/oscar/},\textit{ OPUS corpus} \footnote{http://opus.nlpl.eu/}, \textit{Wikipedia dump} and \textit{Turkish Tree-bank} \footnote{https://ii.metu.edu.tr/metu-corpora-research-group}. The base model has been trained within a cloud environment provided by \textit{Google TensorFlow Research Cloud (TFRC)}. Uncased and cased models with \textit{128K} and \textit{32K} Vocabulary have been successfully trained. In order to load or fine-tune the model, we used \textit{python} and its \textit{transformer} library \footnote{https://huggingface.co/transformers/}, which is built for the NLP community to share, load and extend their models.

The Transformers can learn syntactic segmentation at the lower levels such as \textit{subwords} and then semantic knowledge at the higher levels such as \textit{coreference resolution} as they work on specific language. BERT model applies \textit{WordPiece} tokenization procedure to each token to understand syntactic knowledge. The implementation is simply based on the one from \textit{tensor2tensor}, WordPiece embeddings \cite{wordpiece}. 
%
The \textit{wordpiece} model is generated using a data-driven approach to maximize the language-model likelihood of corpus. When a training corpus is provided, the optimization problem is to select limited wordpieces. Turkish is an agglutinative language and the root word receives many suffix so that incredibly large word could be generated as follows (wikipedia.org)
\begin{verbatim}
 
"muvaffakiyetsizleştiricileştiriveremeyebileceklerimizdenmişsiniz" 

\end{verbatim}

This word is one of the longest words and can be extended more. It is derived from the noun word \textit{"muvaffakiyet"} (success). It means \textit{"As though you are from those whom we may not be able to easily make into a maker of unsuccessful ones"}. The response of wordpiece tokenization procedure to the word is truly successful as follows

\begin{verbatim}
 
['muvaffak',  '##iyet',  '##siz',  '##leş', '##tir',  '##ici',
 '##leş',  '##tir',  '##iver',  '##emeye',  '##bilecekleri', '##mi', '##z', 
 '##den', '##miş', '##siniz', '##cesine',  '##siniz']
\end{verbatim}

It is now transformed into 18 chunks and represented  by (18+2, 32000) tensors in the transformer systems where +2 is needed for [CLS] and [SEP] tokens. Such capacity of the syntactic breakdown helps many models to easily fine-tune downstream tasks. We fine-tuned four models that are now easily accessible via \textit{Hugging-Face platform} \footnote{https://huggingface.co/savasy}.

\subsection{Sentiment Analysis}
The fine-tuned model has been already shared \footnote{https://huggingface.co/savasy/bert-base-turkish-sentiment-cased}.
The dataset for sentiment analysis is obtained by two studies \cite{demirtas}, \cite{mustafasert}. Erkin Demirtas and Mykola Pechenizkiy \cite{demirtas} gathered movie and product reviews from popular e-commerce and social web sites. The set is gathered from a popular Turkish cinema web page \footnote{www.beyazperde.com} with over 5K positive and 5K negative comments in the Turkish language. Reviews are marked in scale from 0 to 5 by the web users. The study considered a review sentiment positive if the rating is equal to or bigger than 4, and negative if it is less or equal to 2 as the study does. The products review dataset is taken from an online e-commerce web page. This benchmark dataset consists of reviews regarding four product categories. The categories of e-commerce dataset are \textit{dvd, books, electronics and kitchen}. Likewise, reviews are marked in the range from 1 to 5. Each category has equally 700 positive and 700 negative reviews, where average rating of negative reviews is 2.27 and of positive reviews is 4.5. The dataset has been evaluated in another study \cite{yildirim} as well. Finally we have organized all the datasets for training, development and test set as described in Table~\ref{senti}

\begin{table}[]
\caption{Dataset for Sentiment Analysis}
\label{senti}
\begin{tabular}{llllll}
\toprule
Sentiment Dataset & \# of total comments &Twitter & E-commerce (with Movie)&  \#Pos & \#Neg \\ 
\midrule
Training Set      & 40830   & 22000 & 18830        & 20340 & 20490 \\ 
Development Set   & 9000   & 4900& 4100        & 4445  & 4556  \\ 
Test Set          & 9000   & 4900 & 4100        & 4391  & 4609  \\ 
\bottomrule
\end{tabular}
\end{table}

The sentiment model was fine-tuned with max-seq-length of 128, learning rate of 2e-5, number of epoch optimized is 3 \footnote{https://huggingface.co/savasy/bert-base-turkish-sentiment-cased}. The performance of the model is reported as shown in Table~\ref{senti2}. We listed other studies performance that used the same e-commerce and Twitter datasets. We only report their best performance for a proper comparison. We got \textit{93.12\%} and \textit{96.69\%} accuracy for e-commerce and Twitter dataset respectively. When we merged two datasets for training and test phase, we got \textit{94.77\%} accuracy . 

\begin{table}[]
\caption{Sentiment Analysis Performance Evaluation}
\label{senti2}

\begin{tabular}{lll}
\toprule
Model          & Dataset         & Results \\
\midrule
Yildirim (2020)     & e-commerce           & 89.26   \\
Demirtas (2013)     & e-commerce          & 85.1    \\
Sert (2017) \cite{mustafasert}   & Twitter              & 80.5    \\
\textit{Our BERT model*} & e-commerce     & \textbf{93.12}  \\
\textit{Our BERT model*} &  Twitter      & \textbf{96.69}  \\
\textit{Our BERT model*} & e-commerce + Twitter      & \textbf{94.77}  \\
\bottomrule
\end{tabular}
\end{table}

\subsection{Named-Entity Recognition}
The fine-tuned model can be found under the link \footnote{https://huggingface.co/savasy/bert-base-turkish-ner-cased}
Named-Entity Recognition aims to capture named entities in a text, such as locations, person. Our NER datasets follow \textit{BIO (Beginning, Inside, Outside)} data format which is a common NER tagging format for tokens. The \textit{B} refers to the first word of an entity and the \textit{I} corresponds to the remaining words of the same entity. Tokens tagged with \textit{O} means are not part of an entity. B and I tags are followed by an entity category such as \textit{B-PER, I-PER, B-ORG}. NER problem is considered a multi-class token classification task. 
There are a couple of named-entity data formats: \textit{Enamex, Timex, Numex}. \textit{Enamex}, which is used in our system, has three entity types: Person, location, organization. NUMEX format is for numerical entities such as money and percent and TIMEX format is used for temporal entities such as date and time. Two different datasets annotated in Enamex format have been used for an evaluation in our fine-tuning phase as shown in Table~\ref{ner}. The first one is Turkish NER dataset obtained from the study \textit{ WikiANN} dataset \cite{wikiann} which includes 282 different language dataset compiled from Wikipedia.  Second one which is also annotated with  POS, LOC, and ORG is shared by NLP community \cite{berturk,yeniterzi2011}

\begin{table*}[]
\caption{NER Dataset}
\label{ner}
\begin{tabular}{llllll}
\toprule
Model      & Dataset    & \# of token & \# PER & \# LOC & \# ORG \\
\midrule
BERT Model & WikiANN    & 340K        & 17.2K  & 19.1K  & 15.8K  \\
BERT Model & Second Dataset & 385K        & 16.6K  & 10.8K  & 10.1K \\
\bottomrule
\end{tabular}
\end{table*}

The initial studies of Turkish NER have started in 1990 \cite{yeniterzi2018}.\textit{ Hidden-markov models, Conditional Random Fields, Bayesian Models and Neural Networks} are among the approaches applied to the NER solution. We listed NER performances of some Turkish studies \cite{yeniterzi2011,gokhan, wikiann,dilekkucuk,hakandemir,gulsen,tur} that especially uses Enamex datasets as shown in Table~\ref{ner2}.  Bert-based fine-tuning model gained comparable results of 95.0 F1 and 95.2 F1 score for two datasets as reported in the table. The study \cite{wikiann} still has the highest score of \textit{96.9 F1} by leveraging cross-lingual language models. They developed a cross-lingual name tagging and linking framework for 282 different languages that exist in Wikipedia. The framework could identify name mentions and assign a coarse-grained type to the mentions. It applies Bi-directional Long Short-Term Memory and Conditional Random Fields (CRFs) network as the underlying learning model for the name tagger for each language. It also links the captured names to an English Knowledge Base.

\begin{table}[]
\caption{Performance Evaluation of NER Models}
\label{ner2}
\begin{tabular}{llll}
\toprule
Model            & Dataset     & F1 Score       &  \\
\midrule
Pan (2017)       &  WikiANN      &\textbf{ 96.9}          &  \\
Kucuk (2014)    &  Twitter data & 72.61         &  \\
Yeniterzi (2018) &  Community Set  & 88.94         &  \\
Demir (2014)    &  by \cite{tur}       & 91.85         &  \\
Seker (2012)     &  by \cite{tur}  & 91.94         &  \\
Seker (2017)     &  by \cite{tur}            & 92.00         & \\
Tur (2003)       &  General News  & 92.73         & \\
Our BERT Model* & WikiANN      & \textbf{95.0} &  \\
Our BERT Model* & Community Set   & \textbf{95.2} &  \\

\bottomrule
\end{tabular}
\end{table}

\subsection{Question Answering}
The fine-tuned model can be found under the link\footnote{https://huggingface.co/savasy/bert-base-turkish-squad} 

Turkish question answering dataset, namely \textit{TQuAD} \footnote{https://github.com/TQuad/turkish-nlp-qa-dataset}, has been prepared in the form of \textit{ Stanford Question Answering Dataset (SQuAD)} which is a well-known English QA dataset. The SQuAD includes questions on a set of Wikipedia articles, where the answer to a question is a span of a text from the corresponding reading passage. While SQuAD 1.0 format does not cover features that the question might be \textit{unanswerable}, this feature was later added on the \textit{SQuAD 2.0} version. The \textit{TQuAD dataset} currently follows SQuAD 1.0. The examples have been manually prepared on \textit{Turkish \& Islamic Science History} within the scope of \textit{Teknofest 2018 Artificial Intelligence competition}, and it has been shared with the NLP community. Training and evaluation set have \textit{681 and 72 titles, 2232 and 275 Paragraphs, and 8308 and 892 Q\&As} respectively. The model has been fine-tuned with the default BERT model settings such as learning rate of 0.00003, maximum sequence length of 384. We observed that model successfully fine-tuned the base pre-trained model with the exact match score of \textit{62.55} and F1 score of \textbf{80.48}. For English Language we can see above \textit{95.0 F1} score and we have enough room for enhancing the Question Answering model by increasing data size and hyper-parameter optimization. Due to the fact that there is no study using SQuAD-like dataset in Turkish, we are not able to compare our model to any Turkish study. 
  
\subsection{Text Classification}
The fine-tuned model can be found under the link \footnote{savasy/bert-turkish-text-classification}.
Text classification downstream task has been fine-tuned with two datasets \textit{TTC-4900} \cite{ttc4900} and \textit{TTC-3600} \cite{ttc3600}. TTC-4900 is originally prepared by \textit{a Turkish NLP Group} \footnote{http://www.kemik.yildiz.edu.tr} and shared with community via repository \footnote{http://www.kaggle.com/savasy/ttc4900}. The data has been already used by many studies \cite{ttc4900}. The dataset has seven coarse-grained categories: \textit{world, economy, culture, health, politics, sport, and technology}. The dataset equally consists of 700 articles under each category. \textit{TTC-3600} dataset has been shared by the study \cite{ttc3600} with a link \footnote{https://github.com/denopas/TTC-3600}. It has six categories; \textit{economy, culture, health, politics, sports, technology}. Each category has 600 examples. The BERT-based model has been fine tuned with these two datasets in the same settings with the original BERT model and reported as shown in Table~\ref{ttc}. As the table indicates, our fine-tuned models surpassed the traditional approaches employed in the other studies \cite{ttc3600,ttc4900} in terms of F1 score by achieving \textbf{93.4} and \textbf{92.2}.

\begin{table*}[]

\caption{Text Classification Performance}
\label{ttc}
\begin{tabular}{llll}
\toprule
Model               & Dataset            & Size & \textit{\textbf{Performance}} \\ 
\midrule
Yildirim (2014)      & TTC 4900 7 Classes  & 4900 & 90.0      \\ 
Kilinc (2017)  & TTC 3600 6 Classes & 3600 & 91.3              \\ 
Our Fine-tuned BERT*  & TTC 4900 7 Classes  & 4900 & \textbf{93.4}         \\ 
Our Fine-tuned BERT*  & TTC 3600 6 Classes  & 3600 & \textbf{92.2}        \\ 
\bottomrule
\end{tabular}
\end{table*}

\section{Final Remarks}

We evaluated four different downstream tasks for the Turkish language. We gathered several datasets used and shared by the community. Other than Question Answering dataset, the remaining three datasets have been already used and evaluated by other studies. We compared our fine-tuned models with those studies who used the same datasets as shown in Table~\ref{senti2}, Table~\ref{ner2}, and Table~\ref{ttc}, which suggests that the BERT-based models outperformed many baseline approaches implemented by other studies. 

We observed some well-known facts that when we transfer the model to another domain, we obtain less performance. For instance, we measure the sentiment prediction scores ranging from 95.0 F1 to 80.0 F1. But when we use the same data from the same domain, the performance improves. As we keep fine-tuning the process fed by that particular domain, the model gets again over 95.0 \% F1 score. The Named-Entity Recognition model follows ENAMEX format where three coarse-grained entity types are considered: PER, LOC, ORG. We want to train NER models for the problems of NUMEX and TIMEX format. Our preliminary studies indicated that we can get high performances around 93 F1 with these formats, which is not reported here. The study \cite{wikiann} has an impressive F1 score of 96.9 and better than our models.

Question Answering model cannot cover unanswerable questions, which means no-answer option is not available and always returns a most likely answer, which is a standard for SQuAD 1.0. We will extend our model to SQuAD 2.0 by adding a "no-answer" option. When comparing our QA study to English counterpart, one can say that both the size of Turkish dataset and the performance of the model needs to be improved. In the last decades, Text Classification problem has been successfully solved by traditional models, TF-IDF, or word embeddings \cite{ttc4900,ttc3600} in Turkish. We showed that the BERT-based model has outperformed the traditional approaches for both seven-class and six-class classification problems. In the future we will fine-tune a model to cover more fine-grained classes.

There exist many other NLU problems to be addressed in the Turkish Language as one of the less studied languages. We also plan to leverage other transformer models such as RoBERTa or ELECTRA  to improve the performance of NLU tasks and solve other types of NLU problems such as anaphora resolution, summarization, relation extraction in Turkish.

\section{Conclusion}
The recent studies showed that transformer-based models got state-of-the-art results in many downstream NLP tasks. In this work, we have leveraged them for four different downstream NLU tasks: Named-Entity Recognition, Sentiment Analysis, Question Answering and Text Classification. The experiments showed that the fine-tuned models significantly outperformed other existing baseline approaches in the Turkish Language as reported in the study. We publicly released these four fine-tuned models in reproducibility and in supporting other researchers. This study is counted to be the first successful attempt to apply the transformer-based model to the Turkish language by fine-tuning the BERT-based model. In the future, we want to leverage other types of language models and cover more challenging down-streams tasks such as coreference resolution or text summarization for the Turkish language.

\bibliographystyle{ACM-Reference-Format}    
\bibliography{sample-base}

\end{document}